\documentclass[10pt,journal,compsoc,onecolumn]{IEEEtran}
\ifCLASSOPTIONcompsoc
  \usepackage[nocompress]{cite}
\else
  \usepackage{cite}
\fi

\ifCLASSINFOpdf
   \usepackage[pdftex]{graphicx}
   \graphicspath{{images/}}

\else

\fi

\usepackage[cmex10]{amsmath}

\PassOptionsToPackage{brazil,portuguese,english}{babel} 
\usepackage{babel}
\usepackage[utf8]{inputenc}
\usepackage{url}
\usepackage[]{algorithm2e}

\usepackage{longtable}
\usepackage{booktabs}
\usepackage{graphicx}
\usepackage{pdflscape}
\usepackage{graphicx}

\usepackage{pgfgantt}
\usepackage{mathtools}
\usepackage{hyperref}
\usepackage{array}
\usepackage{lipsum}

    \usetikzlibrary{shapes,intersections}
\usetikzlibrary {plotmarks,calc,arrows,snakes,automata,backgrounds,petri,positioning,fit}
\usepackage{multirow}
\usepackage{easytable}

\newcolumntype{P}[1]{>{\centering\arraybackslash}p{#1}}
\newcolumntype{M}[1]{>{\centering\arraybackslash}m{#1}}

\tikzset{circarrow/.style={
        *->,
        shorten <=-2pt
    }
}
\tikzset{
    table nodes/.style={
        rectangle,
        draw=black,
        align=center,
        minimum height=7mm,
        text depth=0.5ex,
        text height=2ex,
        inner xsep=0pt,
        outer sep=0pt
    },
    table/.style={
        matrix of nodes,
        row sep=-\pgflinewidth,
        column sep=-\pgflinewidth,
        nodes={
            table nodes
        },
        execute at empty cell={\node[draw=none]{};}
    }
}

\tikzset{place/.style = {circle, draw=blue!50, fill=blue!20, thick, minimum size=0.6cm},
    transition/.style = {rectangle, draw, fill=blue!20, stroke=black, thick, minimum width=3cm,
                        minimum height = 3cm},
    pre/.style =    {<-, semithick},
    post/.style =   {->, semithick}
}

\newcolumntype{M}[1]{>{\centering\arraybackslash}m{#1}}

\usepackage{amsmath}
\usepackage[utf8]{inputenc} 
\usepackage{algorithmic}
\usepackage{multicol}
\usepackage{tikz,pgfplots,pgfplotstable}
\usepackage{setspace}
\usepackage{indentfirst}
\usepackage{enumitem}
\usepackage{multirow}
\usepackage{colortbl}
\definecolor{pblue}{RGB}{0,0,255}
\definecolor{pwhite}{RGB}{255,255,255}
\definecolor{pblack}{RGB}{0,0,0}
\usepackage{here}
\usepackage{pgfplots}
\usepackage{lipsum}
\usepackage{afterpage}
\usepackage{comment}
\usepackage{caption}

\usetikzlibrary{arrows.meta,chains,decorations.pathreplacing,decorations.pathmorphing}
\usetikzlibrary{shapes,intersections,shapes.geometric, arrows, matrix, fit}
\tikzstyle{decision} = [ diamond, draw, fill=blue!20, text width=4.5em, text badly centered, node distance=3cm]
\tikzstyle{block} = [ rectangle, draw, fill=blue!20, text width=5em, text badly centered, rounded corners, minimum height=4em]
\tikzstyle{line} = [ draw, -latex]
\tikzstyle{terminator} = [ draw, ellipse, fill=red!20, node distance=3cm, minimum height=2em]

\usepackage{algorithmic}

\ifCLASSOPTIONcompsoc
  \usepackage[caption=false,font=footnotesize,labelfont=sf,textfont=sf]{subfig}
\else
  \usepackage[caption=false,font=footnotesize]{subfig}
\fi

\begin{document}

\title{Hybrid Model For Word Prediction Using Naive Bayes and Latent Information}

\author{Henrique~X.~Goulart, 
        Mauro~D.~L.~Tosi,
        Daniel~Soares-Gonçalves,
        Rodrigo F. Maia
        and~Guilherme~Wachs-Lopes
}

\IEEEtitleabstractindextext{%
\begin{abstract}
Historically, the Natural Language Processing area has been given too much attention by many researchers. One of the main motivation beyond this interest is related to the word prediction problem, which states that given a set words in a sentence, one can recommend the next word. In literature, this problem is solved by methods based on syntactic or semantic analysis. Solely, each of these analysis cannot achieve practical results for end-user applications. For instance, the Latent Semantic Analysis can handle semantic features of text, but cannot suggest words considering syntactical rules \cite{mikolov2013efficient}. On the other hand, there are models that treat both methods together and achieve state-of-the-art results, e.g. Deep Learning. These models can demand high computational effort, which can make the model infeasible for certain types of applications. With the advance of the technology and mathematical models, it is possible to develop faster systems with more accuracy. This work proposes a hybrid word suggestion model, based on Naive Bayes and Latent Semantic Analysis, considering neighbouring words around unfilled gaps. Results show that this model could achieve 44.2\% of accuracy in the MSR Sentence Completion Challenge.
\end{abstract}

}

\maketitle

\IEEEdisplaynontitleabstractindextext

\IEEEpeerreviewmaketitle

\section{Introduction}

In present days, there is a increasing demand by impaired people to use computer programs, mainly for social interaction. Most of these interactions are made by textual messages, which makes a difficult task for physical impaired people to communicate with others. Thereby, one technique which treats this problem that got attention by researchers is called word prediction.

Word prediction is a word processing feature that aims reduce the number of keystrokes necessary for typing words \cite{aliprandi2008advances}. Usually, these models predict the next word given a set of words based on a context. Because of that, the Natural Language Processing (NLP) area, which performs tasks such word prediction through understanding and interpretation of texts and speeches, became popular.

One of the first NLP approaches in the computer science field is the n-gram model. In this model, each gram is a word in a known set of given words (also known as history). It performs a training in a text database in order to extract information about its language and create a set of features \cite{cavnar1994n}. This approach uses the joint probability table, which increases the hit rate, but that demands too much data knownledge and, many times, it uses unfeasable computer resources \cite{brants2003natural}. For that reason, in order to use less resources, approaches as the Naive Bayes has been developed.

The Naive Bayes is an approach that assumes the conditional independence of its variables.  It requires less memory and processing than approaches that consider this dependence. However, the Naive Bayes accuracy tends to be worst than those because of its knowledge loss during the independence assumption \cite{russell2003}.

The NLP area still has researches for an accurate method that could be implemented and predict in an applicable time. In this context, the latent semantic analysis (LSA) was created. This technique is used to semantically analyze texts through the relationship between the words in different text levels, as phrases, paragraphs, among others \cite{landauer1998introduction}. Besides the high accuracy of the LSA technique, its inferences still are based only on the text frequency, which means that it does not consider word orders and consequently the text syntax.

With the advances of technology, in terms of memory space and processing time, other methods could be developed and consequently implemented, such as the Deep Learning \cite{mikolov2013efficient}. Basically, it consists in a set of artificial neurons which is trained from a database through a multi-layer neural network with the aim of minimize an error function \cite{Goodfellow-et-al-2016}. The Deep Learning process achieves a great precision in sentences completion challenges, however, it has some issues. The computational cost and time to train the amount of data needed is higher compared to other approaches. Beside this, since the learned knowledge is in the weights of neurons connections, it is not possible to interpret what was learned, which means that this is a black box model.

There are many Word prediction models in the NLP area, however, as far as we know none of them can accomplish the task of predict a word with precision in an applicable time, as a human being would do. Therefore, the area still demands development, and a hybrid model might help researches to explore new approaches.

This paper proposes a new hybrid model to predict words using two well-known models:  the Naive Bayes and the LSA. In addition, to this we optimize parameters used to improve the prediction precision through the Gradient Descent technique.

This paper is organized as following. In the next section, will be presented the relevant areas backgrounds. Then, in the third section, the methodology used to the model development and its optimization is explained. Next, in the fourth section, the tests performed to qualify the results that prove the model precision are presented. Then, in the fifth section, the benefits and issues are discussed in the paper conclusion. At last, in the sixth section, some issues that can be improved in future works are presented.

\section{Background}

In NLP area, many models were proposed to predict a word, the Table \ref{Tab_related} shows a list with some related models.

\begin{table}[H]
\begin{center}
	\captionof{table}{Related models.} 
	\scriptsize
    \begin{tabular}{| c | c | c | c | c | c |}
    \hline
    \textbf{Work} & \textbf{Naive Bayes} & \textbf{Other n-gram} & \textbf{LSA} & \textbf{Other model} & \textbf{Year}\\ \hline
	
	Hunnicutt and Carlberger \cite{hunnicutt2001improving} &X&&&&2001 \\ \hline
	Al-Mubaid \cite{al2003context} &X&&&X&2003 \\ \hline
	Al-Mubaid \cite{al2007learning} &X&&&X&2007 \\ \hline
	Aliprandi et al. \cite{aliprandi2008advances} &&&&X&2008 \\ \hline
	Zweig et al. \cite{zweig2012computational} &&X&X&&2012 \\ \hline
	Koutn{\'y}\cite{koutny2012} &&&&X&2012 \\ \hline
	Mikolov et al.\cite{mikolov2013efficient} &&&&X&2013 \\ \hline
	Kleinman et al. \cite{kleinman2015single} &&&&X&2015 \\ \hline
	Spiccia et al. \cite{spiccia2015wordauto} &&&X&&2015 \\ \hline
	Spiccia et al. \cite{spiccia2015wordpos} &&X&X&&2015 \\ \hline
	Luke and Christianson \cite{luke2016limits} &&&X&&2016 \\ \hline
	Cavalieri et al.\cite{cavalieri2016combination} &X&&&&2016 \\ \hline
	
    \end{tabular}
    \label{Tab_related} 
\end{center}
\end{table}

\label{Background}

\subsection{Naive Bayes}
\label{NaiveBayes}

The Naive Bayes is a probabilistic model used in Natural Language Processing (\emph{NLP}) as a n-gram, developed through Bayesian networks that are vastly used in the machine learning area, as stated in \cite{russell2003}. 

Usually, full n-gram models are not feasible to be implemented due to its complexity, because they require a joint probability table. Those tables grow exponentially with the insertion of new variables.

This model states that its variables can be divided in to \emph{cause} and \emph{effect} behaviors; thereby, it can be assumed that the \emph{effects} are conditionally independent between themselves, which reduces the computational cost of the model. This model is mathematically represented in Equation (\ref{Eq_CECompl}), in which, the effect and the cause are represented by $e$ and $c_i$ respectively. Thus, using this model it is not necessary to build a joint probability table.
\begin{equation}
    P(e|c_1, \cdots, c_n) = \frac {P(e) \prod_{i=1}^{n} P(c_i|e)}{\gamma}
\label{Eq_CECompl}
\end{equation}

The normalization factor $\gamma$ is calculated by the Equation (\ref{Eq_gamma}), in which $k$ represents the number of words learned in the train.

\begin{equation}
\label{Eq_gamma}
\gamma = \sum_{j=1}^{k} P(e_j| c_1, ..., c_n)
\end{equation}  

\subsection{Latent Semantic Analysis}
\label{Sec_BackLSA}

The latent semantic analysis (LSA) has been vastly used in systems that analyze textual contents \cite{zupanc2017automated}, since it performs a comparison between words to infer.

The LSA can use any cohesive textual level, such as phrases, paragraphs, entire documents in its traning set \cite{coccaro1998towards}. The information (text) of this textual levels contains the semantic relationship between the words in it. To store these relations, a table is constructed (\emph{relationship table}) containing the frequency ($f_{i,j}$) that any word $i$ appeared in a textual level $j$.

With the \emph{relationship table}, comparisons between words can be performed to discover similar words. Therefore, it is necessary to compute the distance between the \emph{relationship table} lines using some metrics \cite{zhila2013combining}. 

Thereby, the LSA is a technique that is used to extract and infer the words contextual usage relationship in a vector space, as stated in \cite{landauer1998introduction}.

A common problem of the LSA is the dimension of the \emph{relationship table}, that is usually sparse and consumes unnecessary memory space. A solution used in the area for this issue \cite{spiccia2015wordauto}\cite{coccaro1998towards} is the usage of the Singular Value Decomposition (SVD) technique. With it, it is possible to reduce the \emph{relationship table} size without losing all the original information, keeping it dense. In this work, the reduced table is noted as \emph{Semantic Reduced Table} ($SRT$).

\subsection{Gradient Descent}

The gradient descent is a mathematical optimization algorithm used to modify variables values based on their contribution to minimize an error function \cite{witten2016data}. Often, this technique is used in prediction models such as: neural networks, deep learning, bayesian learning, among many others.

To perform the optimization, the error function has to be differentiable, because the gradient descent algorithm uses derivatives. The most used error function is the squared-error \cite{witten2016data}, modeled as Equation (\ref{Eq_erroGradiente}), in which $f(x)$ is the model prediction, $y$ is the ideal value of the prediction (usually 0 or 1) and the number $\frac{1}{2}$ is used to simplify the further derivatives calculus, as it will be dropped in the next processes.

\begin{equation}
\label{Eq_erroGradiente}
E = \frac{1}{2}(y - f(x))^2
\end{equation}

The variables values changes are proportional to the error function gradient \cite{homod2012gradient}. Thereby, to minimize the error, the variables have to be updated proportionally in the opposite direction of gradient function. To smooth the optimization updates, a factor $\eta$ is often used to control and keep the current knowledge of the model, compared with the recent information acquired.

\section{Proposed Model}

\begin{figure}[h]
\centering
	\caption{Proposed model}
	\includegraphics[width=0.7\linewidth]{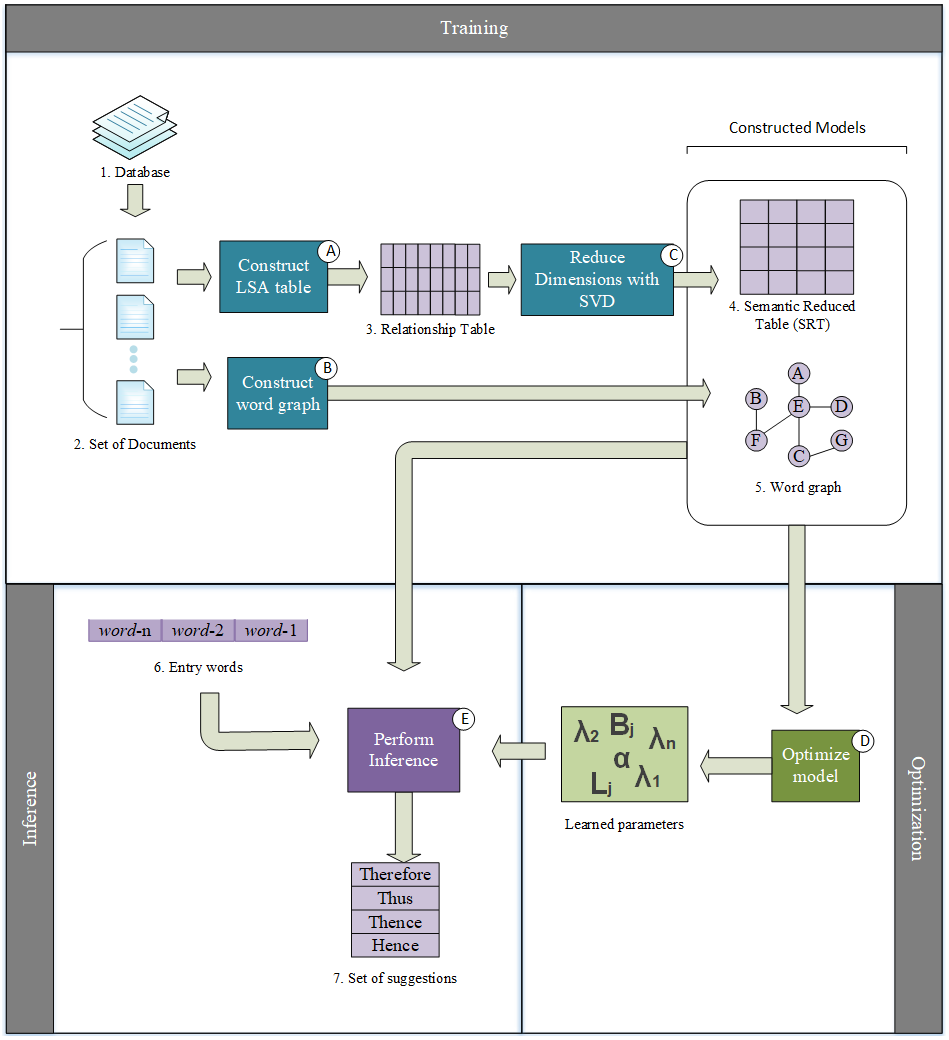}
	\label{fig:funcionamentoSistemaServidor}
\end{figure}

This paper proposes a hybrid word prediction model, that performs inferences based on Naive Bayes and Latent Semantic Analysis (LSA) theories. The methodology used to develop the proposed model is divided into three stages: Training, Optimization and Inferences, illustrated in Figure \ref{fig:funcionamentoSistemaServidor}. Those steps are described and analyzed in this section.

\subsection{Training}

To perform probabilistic inferences, it is necessary to create and train the LSA and Naive Bayes sub-models. These sub-models are trained in order to represent two types of information. The former, Naive Bayes, is used to store co-occurrences patterns of words; the later, LSA, is used to model the language semantics.

In order to train this model, a textual database is required. Therefore, the Project Gutenberg database \cite{projgten} was used to accomplish this pre-requisite as it is used vastly in literature such as in \cite{spiccia2015wordauto}\cite{gubbins2013dependency}\cite{zweig2012computational}. A set composed by 522 19th Century literature books was used in this paper.

\subsubsection{Naive Bayes Graph}
\label{sec:naive}

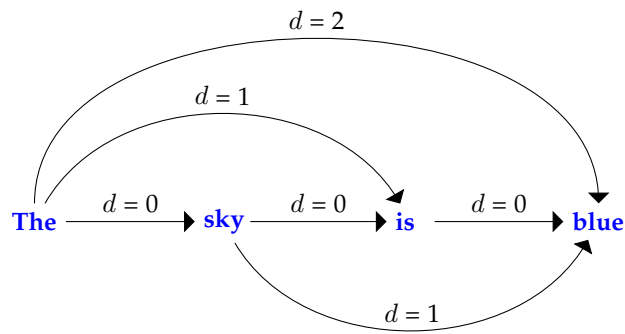
\begin{figure}[h!]
    \caption{Words Distance ($d$) for phrase: "The sky is blue".}
\centering
\begin{tikzpicture}
    \node[text width=0.6cm] (n0) at (0,0) {\color{blue}{\textbf{The}}};
    \node[text width=0.5cm] (n1) at (2.5,0) {\color{blue}{\textbf{sky}}}; 
    \node[text width=0.4cm] (n2) at (5,0) {\color{blue}{\textbf{is}}};
    \node[text width=0.7cm] (n3) at (7.5,0) {\color{blue}{\textbf{blue}}};

    \path (n0) edge [bend right, draw, -triangle 90, out=90, in=90] node[sloped, anchor=center, above,align=center]{$d$ = 2}(n3);
    \path (n0) edge [bend right, draw, -triangle 90, out=60, in=120] node[sloped, anchor=center, above,align=center]{$d$ = 1}(n2);
    \path (n0) edge [bend right, draw, -triangle 90, out=0, in=180] node[sloped, anchor=center, above,align=center]{$d$ = 0}(n1);blue is a color

    \path (n1) edge [bend left, draw, -triangle 90, out=300, in=240] node[sloped, anchor=center, above,align=center]{$d$ = 1}(n3);
    \path (n1) edge [bend left, draw, -triangle 90, out=0, in=180] node[sloped, anchor=center, above,align=center]{$d$ = 0}(n2);

    \path (n2) edge [bend left, draw, -triangle 90, out=0, in=180] node[sloped, anchor=center, above,align=center]{$d$ = 0}(n3);

\end{tikzpicture}
\label{fig:wordsDistance}
\end{figure}

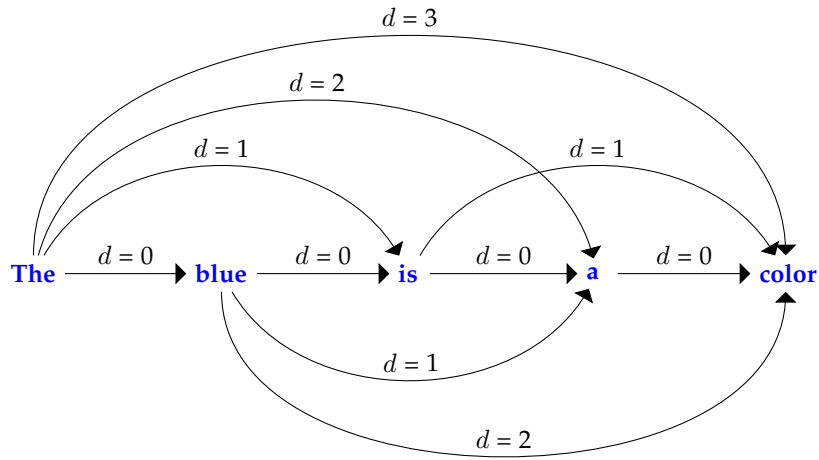
\begin{figure}[h!]
    \caption{Words Distance ($d$) for phrase: "The blue is a color".}
\centering
\begin{tikzpicture}
    \node[text width=0.6cm] (n0) at (0,0) {\color{blue}{\textbf{The}}};
    \node[text width=0.7cm] (n1) at (2.5,0) {\color{blue}{\textbf{blue}}}; 
    \node[text width=0.3cm] (n2) at (5,0) {\color{blue}{\textbf{is}}};
    \node[text width=0.3cm] (n3) at (7.5,0) {\color{blue}{\textbf{a}}};
    \node[text width=0.7cm] (n4) at (10,0) {\color{blue}{\textbf{color}}};

	\path (n0) edge [bend right, draw, -triangle 90, out=90, in=90] node[sloped, anchor=center, above,align=center]{$d$ = 3}(n4);
    \path (n0) edge [bend right, draw, -triangle 90, out=75, in=105] node[sloped, anchor=center, above,align=center]{$d$ = 2}(n3);
    \path (n0) edge [bend right, draw, -triangle 90, out=60, in=120] node[sloped, anchor=center, above,align=center]{$d$ = 1}(n2);
    \path (n0) edge [bend right, draw, -triangle 90, out=0, in=180] node[sloped, anchor=center, above,align=center]{$d$ = 0}(n1);

    \path (n1) edge [bend left, draw, -triangle 90, out=300, in=240] node[sloped, anchor=center, above,align=center]{$d$ = 1}(n3);
    \path (n1) edge [bend left, draw, -triangle 90, out=0, in=180] node[sloped, anchor=center, above,align=center]{$d$ = 0}(n2);
    \path (n1) edge [bend left, draw, -triangle 90, out=270, in=270] node[sloped, anchor=center, above,align=center]{$d$ = 2}(n4);

    \path (n2) edge [bend left, draw, -triangle 90, out=0, in=180] node[sloped, anchor=center, above,align=center]{$d$ = 0}(n3);
    \path (n2) edge [bend right, draw, -triangle 90, out=60, in=120] node[sloped, anchor=center, above,align=center]{$d$ = 1}(n4);
    
    \path (n3) edge [bend left, draw, -triangle 90, out=0, in=180] node[sloped, anchor=center, above,align=center]{$d$ = 0}(n4);

\end{tikzpicture}
\label{fig:wordsDistance2}
\end{figure}

The co-occurrences patterns which represents the Naive Bayes network can be stored in a set of graphs $G = (g_0, g_1, ..., g_{d-1})$, where nodes represent words and edges represent the number of times that each pair of words co-occurred in a same textual level. Each graph $g_i$ with $0 <= i <= {n-1}$ represents the connections of the words that appear in the text with $i$ words between them. This distances are illustrated in the Figures \ref{fig:wordsDistance} and \ref{fig:wordsDistance2}, taking the phrases "The sky is blue" and "The blue is a color" as example. 

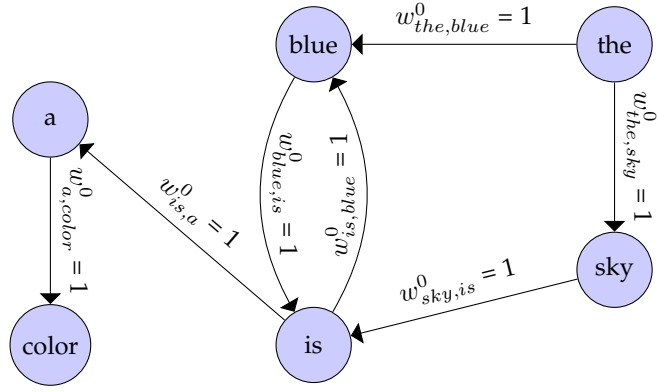
\begin{figure}[h!]
\centering
    \caption{Graph 0 with weights ($w$).}
\begin{tikzpicture}
	
	\draw (2,1) node (n0) [draw, circle,fill=blue!20, minimum size=1cm] {the};
	\draw (2,-2) node (n1) [draw, circle,fill=blue!20, minimum size=1cm] {sky};
	\draw (-2,-3) node (n2) [draw, circle,fill=blue!20, minimum size=1cm] {is};
	\draw (-2,1) node (n3) [draw, circle,fill=blue!20, minimum size=1cm] {blue};
	\draw (-5.5,0) node (n4) [draw, circle,fill=blue!20, minimum size=1cm] {a};
	\draw (-5.5,-3) node (n5) [draw, circle,fill=blue!20, minimum size=1cm] {color};
    
	\path[draw, -triangle 90] (n0) -> (n1) node [midway, above, sloped] () {$w^0_{the,sky}$ = 1};
   	\path[draw, -triangle 90] (n1) -> (n2) node [midway, above, sloped] () {$w^0_{sky,is}$ = 1};
   	\path[draw, -triangle 90] (n0) -> (n3) node [midway, above, sloped] () {$w^0_{the,blue}$ = 1};
    \path[draw, -triangle 90] (n2) -> (n4) node [midway, above, sloped] () {$w^0_{is,a}$ = 1};
	\path[draw, -triangle 90] (n4) -> (n5) node [midway, above, sloped] () {$w^0_{a,color}$ = 1};
    \path (n2) edge [bend right, draw, -triangle 90] node[sloped, anchor=center, above,align=center]{$w^0_{is,blue}$ = 1}(n3);
    \path (n3) edge [bend right, draw, -triangle 90] node[sloped, anchor=center, above,align=center]{$w^0_{blue,is}$ = 1}(n2); 
\end{tikzpicture}

\label{fig:distanceGraph0}
\end{figure}

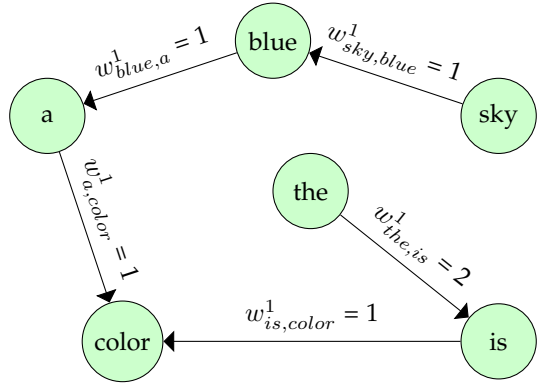
\begin{figure}[h!]
\centering
    \caption{Graph 1 with weights ($w$).}
\begin{tikzpicture}
	
	\draw (-1.5,-1) node (n0) [draw, circle,fill=green!20, minimum size=1cm] {the};
	\draw (1,0) node (n1) [draw, circle,fill=green!20, minimum size=1cm] {sky};
	\draw (1,-3) node (n2) [draw, circle,fill=green!20, minimum size=1cm] {is};
	\draw (-2,1) node (n3) [draw, circle,fill=green!20, minimum size=1cm] {blue};
	\draw (-5,0) node (n4) [draw, circle,fill=green!20, minimum size=1cm] {a};
	\draw (-4,-3) node (n5) [draw, circle,fill=green!20, minimum size=1cm] {color};
    
	\path[draw, -triangle 90] (n0) -> (n2) node [midway, above, sloped] () {$w^1_{the,is}$ = 2};
   	\path[draw, -triangle 90] (n1) -> (n3) node [midway, above, sloped] () {$w^1_{sky,blue}$ = 1};
   	\path[draw, -triangle 90] (n3) -> (n4) node [midway, above, sloped] () {$w^1_{blue,a}$ = 1};
    \path[draw, -triangle 90] (n2) -> (n5) node [midway, above, sloped] () {$w^1_{is,color}$ = 1};
	\path[draw, -triangle 90] (n4) -> (n5) node [midway, above, sloped] () {$w^1_{a,color}$ = 1};i
\end{tikzpicture}

\label{fig:distanceGraph1}
\end{figure}

The $G$ set is constructed from many different graphs $g_i$ for all $i$. Each $g_i$ is constructed taking every co-occurrences of distance $i$ in a text. The weight of edges will be higher as many co-occurrences appear in it, this is represented by $w^d_{i,j}$, where $d$ is the graph and $i,j$ represents the edge which connects words $i$ and $j$; the value of these edges is the number of times that words $i$ and $j$ occurred in the database, establishing this approach as a frequentist model. Take as example the Figures \ref{fig:distanceGraph0} and \ref{fig:distanceGraph1} that represents the $g_0$ and $g_1$ respectively which was constructed using the same sentences used previously: "The sky is blue" and "The blue is a color".

\subsubsection{Latent Semantic Analysis (LSA)}

This step consists of building the \emph{relationship table}. Firstly, it is necessary to adopt the textual level to be used in the LSA. In this paper, only sentences with more than 4 nonstop-words are used. This measure was established to gather only the most semantic relevant phrases \footnote{The source-code used as part of implementation of this sub-model can be found in following link: https://github.com/chiawen/sentence-completion}.

To construct the \emph{relationship table}, each different word in the database is represented as a row and each textual level as a column. The value of each table cell $c_{word,t}$ is defined by the number of times that the word is present in the textual level $t$.

As stated before in Section \ref{Sec_BackLSA}, the $STR$ is obtained from \emph{relationship table}, where the inferences can be performed using acceptable computer resources. 

\subsection{Inferences}

Based on the $n$ previously used words in a text, inferences can be performed in the constructed models to predict a word usage probability in the analyzed context. This section will describe these inferences using the notation $prev_{0:{n-1}}$ to represent all the $n$ previous words.

\subsubsection{Naive Bayes inference}
\label{Sec_nbi}

The bayesian inferences are computed to establish the $P(sugg_j | prev_0, ..., prev_{n-1})$, in which $sugg_j$ is a possible word recommendation (suggestion) from $prev_0, ..., prev_{n-1}$ history.

Thereby, the Equation (\ref{Eq_recomendacaoBayes}) represents the probability of the $sugg_j$ be used just after the other $prev_{0:n-1}$ words in the same sentence. This probability is represented by the notation $B_j$. Furthermore, each $P^i(prev_i|sugg_j)$ is a probability created by a normalization of $w^d_{prev_i,sugg_j}$ (explained in Section \ref{sec:naive}) illustrated in Equation (\ref{Eq_normProb}). The prior probability $P(sugg_j)$ was gathered directly from database, which is represented in Equation (\ref{Eq_normInd}).

\begin{equation}
\label{Eq_recomendacaoBayes}
B_j = \frac{P(sugg_j)\prod_{i=0}^{n-1} P^i(prev_i|sugg_j)}{\gamma}
\end{equation}

\begin{equation}
\label{Eq_normProb}
P^d(i|j) = \frac{w^d_{i,j}}{\sum_{j} w^d_{i,j}}
\end{equation}

\begin{equation}
\label{Eq_normInd}
P(i) = \frac{\textnormal{number of occurrences of }i}{\textnormal{total of word occurrences in database}}
\end{equation}

Therefore, all the previously used words have the same relevance to the model, as the Naive Bayes considers the events independently, which decreases the model precision, because as explained previously, the Naive Bayes not considers all probabilities involved in inference. In order to minimize this issue, variables responsible to weight this relevance were inserted in the model, noted as $\lambda_i$, in which $i$ refers to the distance between the weighted and the analyzed words.

The equation $B_j$ now represents the weighted model, Figure \ref{Eq_bayesLamdas}, in which $P(prev_0|sugg_j)$ is inversely weighted by the $\lambda_{1,...,n-1}$. Thereby, the value of $\lambda_{1,...,n-1}$ are normalized between themselves, and the equation energy is preserved.        

\begin{equation}
\label{Eq_bayesLamdas}
B_j = \frac{P(sugg_j)P(prev_0|sugg_j)^{\frac{1}{\lambda_1 * ... * \lambda_{n-1}}}\prod_{i=1}^{n-1} P^i(prev_i|sugg_j)^{\lambda_i}}{\gamma}
\end{equation}

\subsubsection{Latent Information Inference}

In order to use the knowledge of the $prev_{0:n-1}$ words, $n$ inferences have to be computed in the $SRT$, which directly impacts the computational cost of the model. Those are executed by calculating the semantic similarity of each $prev_i$ word, $0 <= i < n-1$, with all $SRT$ words in vocabulary. Those similarities are calculated by the second norm distance, in which, each vector represents a word.



Therefore, each of the $n$ $SRT$ cells represents the distance between each candidate word $j$ and $prev_i$. Thus, to obtain the similarity of $j$ to all $prev$ words, the inverse of the distance vector values are summed, as the distance is inversely proportional to the semantic similarity of the words. Thus, the semantic distance $d_j$ between $j$ and $prev_{0:n-1}$ is represented by Equation (\ref{Eq_distTotalLSA}).

\begin{equation}
d_j = \sum_{i=0}^{n-1} \frac{1}{ \| j - i \|_2 + 1}
\label{Eq_distTotalLSA}
\end{equation}

After calculating $d_j$, it is necessary to normalize the results based in $n$, as in Equation (\ref{Eq_lsa}). Therefore, $L_j$ represents the normalized semantic similarity of the word $j$ based on the $n$ previous known words.

\begin{equation}
L_j = \frac{1}{n}*d_js
\label{Eq_lsa}
\end{equation}

\subsubsection{Hybrid Inference}
\label{sec:hybridModel}

The trained networks achieved through the Naive Bayes and LSA models, and their respective inferences, Equations (\ref{Eq_bayesLamdas}) and (\ref{Eq_lsa}), output different pattern results, which affects the hybrid model development. The probability variance of the Naive Bayes inferences is much larger than the LSA ones. 

To successfully merge the models, it is necessary to establish an output pattern to equalized them. Therefore, the probability values of the next word to be inferred, from both models, are sorted in ascending order crescent in two vectors. Thereby, those probabilities are replaced by their vector index and normalized by the sum of all indexes, as illustrated below.

\begin{table}[H]
\centering
\begin{tabular}{|l|l|l|l|l|}
\hline
Probabilities & 0.5 & 0.15 & 0.3 & 0.05 \\ \hline
\end{tabular}
\end{table}

\begin{table}[H]
\centering
\begin{tabular}{l}
$\Downarrow$
\end{tabular}
\end{table}

\begin{table}[H]
\centering
\begin{tabular}{|l|l|l|l|l|}
\hline
Probalities Indexes & 4 & 2 & 3 & 1 \\ \hline
\end{tabular}
\end{table}

\begin{table}[H]
\centering
\begin{tabular}{l}
$\Downarrow$
\end{tabular}
\end{table}

\begin{table}[H]
\centering
\begin{tabular}{|l|l|l|l|l|}
\hline
Equalized Probabilities & $\frac{4}{10}$ & $\frac{2}{10}$ & $\frac{3}{10}$ & $\frac{1}{10}$\\ \hline
\end{tabular}
\end{table}

Thus, the Naive Bayes and LSA inferences are weighted, which creates the hybrid model, that is represented in Equation (\ref{Eq_misto}).

\begin{equation}
\theta_j = \alpha B_j + (1 - \alpha) L_j
\label{Eq_misto}
\end{equation}

The proposed model uses a weighting constant $alpha$, that aims to improve the inferences precision. This constant represents the Naive Bayes percentage relevance in comparison with the LSA. For example, if $\alpha = 1$ all the inference will be done through the Naive Bayes else, if $\alpha = 0$, through the LSA.

\subsection{Optimization}

Once Naive Bayes and LSA networks are trained, the inferences can be performed. However, those models provide a different set of probabilities ($B_j$ and $L_j$). Considering the hybrid model of Equation (\ref{Eq_misto}), there is a set parameters that can be optimized in order to provide an ideal weight between the models, improving the recommendation precision. Moreover, the values of $\lambda$ parameters can also be optimized, achieving a better weight between the partial probabilities of Naive Bayes.

Therefore, optimization through supervised training was performed. The set of optimized parameters was composed by $\alpha$, that represents the relevant percentage of the Naive Bayes compared to the LSA; and $\lambda_1,...,\lambda_{n-1}$, that express the weight/relevance relation between the words distances in the Naive Bayes inference.

As the inference goal is to predict the correct word with the maximum precision, it is desired that the correct inferred value be as close as possible to 100\%. Thus, the Equation (\ref{Eq_Erro}) can be determined, in which $E$ represents the system error. The optimization proposed in this work is performed using only the probability output of the correct suggestion $sugg_c$ (thus, $B_c$ and $L_c$ are used for optimization).


\begin{equation}
E = \frac{1}{2}(1 - \theta_{j=c})^{2} ~~~,~\mbox{for}~c~=~\mbox{``Correct suggestion''}
\label{Eq_Erro}
\end{equation}

After expanding the equation of error, the Equation (\ref{Eq_erro_Extendido}) is obtained.

\begin{equation}
E = \frac{1}{2}\left[1 - (\alpha B_c + (1 - \alpha) L_c)\right]^{2} ~~~,~\mbox{for}~c~=~\mbox{``Correct suggestion''}
\label{Eq_erro_Extendido}
\end{equation} 

\subsubsection{Alpha ($\alpha$)}

The gradient descent optimization technique was used to find the ideal value of $\alpha$. In this case, its value was updated using the error function through an iterative process. Using Equation (\ref{Eq_erro_Extendido}), the $\alpha$ update is performed using its derivative over the $\alpha$ itself represented by Equation (\ref{Eq_derivadaAlpha}). Therefore, the error value will tend to decrease.

\begin{equation}
\frac{\partial E}{\partial \alpha} = - (1 - \theta_c) * (B_c - L_c)
\label{Eq_derivadaAlpha}
\end{equation}

Equations (\ref{Eq_alphaLinha}) and (\ref{Eq_alphaLinhaSimp}) add a new term to the optimization, in which $\eta_{\alpha}$ represents the learning rate of each iteration, regarding the current knowledge of the model. 

\begin{equation}
\alpha ' = \alpha - \left(\frac{\partial E}{\partial \alpha}\right) * \eta_\alpha
\label{Eq_alphaLinha}
\end{equation}

\begin{equation}
\alpha ' = \alpha + (1 - \theta_c) * (B_c - L_c) * \eta_\alpha
\label{Eq_alphaLinhaSimp}
\end{equation}

\subsubsection{Lambda ($\lambda$)}

The $\lambda_1, ..., \lambda_{n-1}$ values were optimized similarly to the $\alpha$ one. A supervised training was performed with thousands of iterations in which, each $\lambda_1, ..., \lambda_{n-1}$ was updated regarding it respective error in each iteration.

The error, Equation (\ref{Eq_erro_Extendido}), can be expanded to Equation (\ref{Eq_erroLambda}), in which the equalized $P(prev_i|sugg_j)$, as explained in Section \ref{sec:hybridModel}, and the $P(sugg_j)$ have been replaced by $t_i$ and $t$ respectively.

\begin{equation}
E = \frac{1}{2} \left(1 - \left(\frac{\alpha * t*...*t_2^{\lambda_2}*t_1^{\lambda_1}*t_0^{\frac{1}{\lambda_1*...\lambda_{n-1}}}}{\gamma}+ (1 - \alpha) L_j\right)\right)^{2}
\label{Eq_erroLambda}
\end{equation}

Therefore, with the derivative of the $E$ over each $\lambda$, an individual error rate value is obtained. The Equations (\ref{Eq_derLambda1}) and (\ref{Eq_derLambda2}) represent the derivative of the $E$ regarding $\lambda_1$ and $\lambda_2$ respectively, in a model that considers only three previous words.

\begin{equation}
\frac{\partial E}{\partial \lambda_1} = -\alpha  (1 - \theta)	\left[\frac{t*t_2^{\lambda_2}*t_1^{\lambda_1}*t_0^{\frac{1}{\lambda_1 * \lambda_2}}}{\gamma}\left(ln(t_1)-\frac{ln(t_0)}{\lambda_1^2 * \lambda_2}\right)\right]
\label{Eq_derLambda1}
\end{equation}


\begin{equation}
\frac{\partial E}{\partial \lambda_2} = -\alpha  (1 - \theta) \left[\frac{t*t_2^{\lambda_2}*t_1^{\lambda_1}*t_0^{\frac{1}{\lambda_1 * \lambda_2}}}{\gamma}\left(ln(t_2)-\frac{ln(t_0)}{\lambda_1 * \lambda_2^2}\right)\right]
\label{Eq_derLambda2}
\end{equation}

With the derivative of the $E$ over each $\lambda_x$, $1 <= x <= n-1$, the $\lambda_x$ value is updated over each iteration, Equation (\ref{Eq_LambdaLinha}), in which $\eta_{\lambda}$ represents the learning rate of each iteration, regarding the current knowledge of the model.

\begin{equation}
\lambda_{x}' = \lambda_{x} - \frac{\partial E}{\partial \lambda_x} * \eta_\lambda
\label{Eq_LambdaLinha}
\end{equation}

\section{Results}

In order to validate the proposed model, it was applied the experiment MSR Sentence Completion Challenge \cite{zweig2011microsoft}, this experiment consists of the 1040 sentences with a missing word. Moreover, it is provided 5 options to choose a word for complete the sentence.

\begin{figure}[h]
\centering
	\caption{Example of the bayesian inference for MSR Sentence Completion Challenge.}
	\label{Fig_mirror}
\begin{tikzpicture}[->,node distance=1.3cm,>=stealth',bend angle=40,auto,
  place/.style={rectangle,thick,draw=black,fill=cyan!20,minimum size=10mm},
  red place/.style={place,draw=red!75,fill=red!20}
  every label/.style={red},
  every node/.style={scale=1},
  dots/.style={fill=black,circle,inner sep=2pt},
  initial text={}, scale=0.9, every node/.style={scale=0.9}]

  \node [place, fill=red!20] (inf) {?};

  \node [place,left=5cm of inf, fill=green!20] (ant2) {Lorem};
  \node [place,left=3.5cm of inf, fill=green!20] (ant1) {ipsum};
  \node [place,left=2cm of inf, fill=green!20] (ant0) {dolor};

  \draw (ant2) to[in=-90,out=-90] node[below,align=center] {Distance 2} (inf);
  \draw (ant1) to[in=135,out=45] node[above,align=center] {Distance 1} (inf);
  \draw (ant0) to[in=180,out=0] node[below,align=center] {Distance 0} (inf);

  \node [place,right=2cm of inf] (pro0) {amet};
  \node [place,right=3.5cm of inf] (pro1) {vitae};
  \node [place,right=5cm of inf] (pro2) {elit};

  \draw (pro2) to[in=-90,out=-90] node[below,align=center] {Distance -2} (inf);
  \draw (pro1) to[in=45,out=135] node[above,align=center] {Distance -1} (inf);
  \draw (pro0) to[in=0,out=180] node[below,align=center] {Distance 0} (inf);
\end{tikzpicture}

\end{figure}
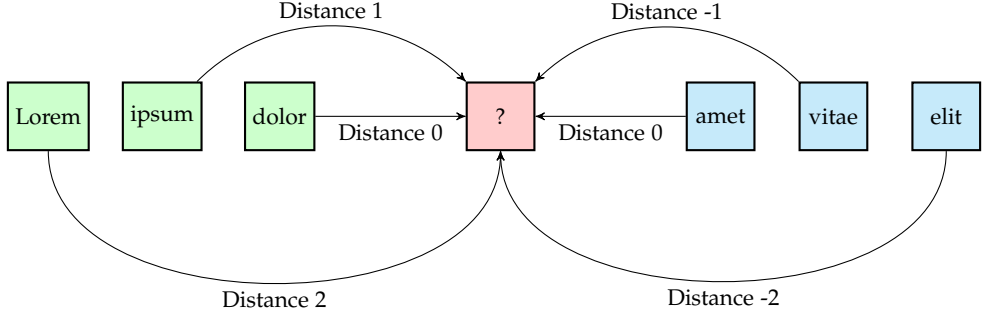

In the MSR Sentence Completion Challenge, it is possible that the missing word is in the middle of sentence. For this case, the bayesian model also considers the posterior words from the gap. This case is illustrated in Figure \ref{Fig_mirror} using as example the following sentence: \emph{Lorem ipsum dolor \_\_\_ amet vitae elit}. Thereby, the LSA also consider all words it was considered by Naive Bayes.

Thus, the Naive Bayes inference was described by Equation (\ref{Eq_lambfull}).

\begin{equation}
B_j = \frac{t*b_2^{\lambda_{b2}}*b_1^{\lambda_{b1}}*b_0^{\frac{1}{\lambda_{b1}*\lambda_{b2}}} *a_0^{\frac{1}{\lambda_{a-1}*\lambda_{a-2}}}*a_{-1}^{\lambda_{a-1}}*a_{-2}^{\lambda_{a-2}}}{\gamma}
\label{Eq_lambfull}
\end{equation}

\begin{figure}[h]
\centering
	\caption{Experiment configurations.}
	\label{Fig_boxTest}
	\scriptsize

\begin{tikzpicture}[
node distance=0pt,
scale=0.7, every node/.style={scale=1},
 start chain = A going right,
arrow/.style = {draw=#1,-{Stealth[]},
                shorten >=1mm, shorten <=1mm}, 
arrow/.default = black,
    X/.style = {rectangle, draw,
                minimum width=2ex, minimum height=3ex,
                outer sep=0pt, on chain},
    B/.style = {decorate,
                decoration={brace, amplitude=5pt,
                pre=moveto,pre length=1pt,post=moveto,post length=1pt,
                raise=1mm,
                            #1}, 
                thick},
   B/.default = mirror, 
                    ]

\matrix (MR) [matrix of nodes,
              nodes=draw, dashed, row sep=1mm,
              row 1 column 1/.style={draw=red, nodes={fill=red!20}}]
{   Group 1\\
    Group 2\\
    Grupo 3\\
    Group 4\\
    Group 5\\
};
\draw   (MR.north -| MR-1-1.north west) -|
        (MR.south west) --
        (MR.south -| MR-3-1.south west)
        (MR.north -| MR-1-1.north east) -|
        (MR.south east) --
        (MR.south -| MR-3-1.south east)
        ;

\node [above=3mm of MR] (inf) {Config. 1};

\matrix (MS) [matrix of nodes,
              nodes=draw, dashed, row sep=1mm,
              row 2 column 1/.style={draw=red, nodes={fill=red!20}},
              right=5mm of MR]
{   Group 1\\
    Group 2\\
    Group 3\\
    Group 4\\
    Group 5\\
};
\draw   (MS.north -| MS-1-1.north west) -|
        (MS.south west) --
        (MS.south -| MS-3-1.south west)
        (MS.north -| MS-1-1.north east) -|
        (MS.south east) --
        (MS.south -| MS-3-1.south east)
        ;

\node [above=3mm of MS] (inf) {Config. 2};

\matrix (MT) [matrix of nodes,
              nodes=draw, dashed, row sep=1mm,
              row 3 column 1/.style={draw=red, nodes={fill=red!20}},
              right=5mm of MS]
{   Group 1\\
    Group 2\\
    Group 3\\
    Group 4\\
    Group 5\\
};
\draw   (MT.north -| MT-1-1.north west) -|
        (MT.south west) --
        (MT.south -| MT-3-1.south west)
        (MT.north -| MT-1-1.north east) -|
        (MT.south east) --
        (MT.south -| MT-3-1.south east)
        ;

\node [above=3mm of MT] (inf) {Config. 3};

\matrix (MU) [matrix of nodes,
              nodes=draw, dashed, row sep=1mm,
              row 4 column 1/.style={draw=red, nodes={fill=red!20}},
              right=5mm of MT]
{   Group 1\\
    Group 2\\
    Group 3\\
    Group 4\\
    Group 5\\
};
\draw   (MU.north -| MU-1-1.north west) -|
        (MU.south west) --
        (MU.south -| MU-3-1.south west)
        (MU.north -| MU-1-1.north east) -|
        (MU.south east) --
        (MU.south -| MU-3-1.south east)
        ;

\node [above=3mm of MU] (inf) {Config. 4};

\matrix (MV) [matrix of nodes,
              nodes=draw, dashed, row sep=1mm,
              row 5 column 1/.style={draw=red, nodes={fill=red!20}},
              right=5mm of MU]
{   Group 1\\
    Group 2\\
    Group 3\\
    Group 4\\
    Group 5\\
};
\draw   (MV.north -| MV-1-1.north west) -|
        (MV.south west) --
        (MV.south -| MV-3-1.south west)
        (MV.north -| MV-1-1.north east) -|
        (MV.south east) --
        (MV.south -| MV-3-1.south east)
        ;

\node [above=3mm of MV] (inf) {Config. 5};

\matrix (leg) [matrix of nodes,
              nodes=draw, dashed, row sep=2mm,
              row 2 column 1/.style={draw=red, nodes={fill=red!20}},
              right=10mm of inf]
{   ~~~~\\
    ~~~~\\
};

\node [right=1mm of leg-1-1] (leg1) {Optimization};
\node [right=1mm of leg-2-1] (leg2) {Test};

\matrix (legg) [matrix of nodes, row sep=2mm,nodes=draw,
				column 1/.style={draw=white},
              right=17mm of leg]
{   ~~~~\\
    ~~~~\\
};

\draw   (leg.north west) -|
        (leg.south west)
        (legg.north west) -|
        (legg.south west)
        (legg.north west) -|
        (leg.north west)
        (legg.south west) -|
        (leg.south west)
        ;
        

\end{tikzpicture}

\end{figure}
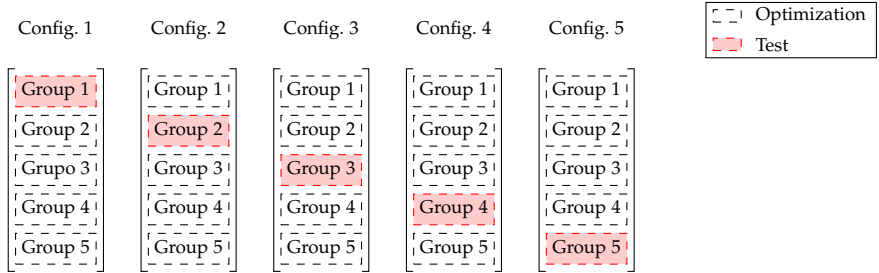

To guarantee cross-validation on experiments, the MSR Sentence Completion Challenge was separated into 5 groups. Therefore, one of these groups was used for test and the others for optimization, this is illustrated in Figure \ref{Fig_boxTest}. Furthermore, is possible to have 5 different and independent configurations for experiments.

\subsection{Error Variation}

In order to validate the proposed optimization, it was created a scenario where the parameters $\lambda$ and $\alpha$ were initialized randomly in the range 0 and 1 for each configuration. Furthermore, the optimization groups were disposed in a cyclic loop (i.e. epochs), thus extending the optimization. In this scenario, it was used a 3-gram history.

\begin{figure}
\centering
	\caption{Variation of error throughout the epochs.}
	\includegraphics[width=0.7\linewidth]{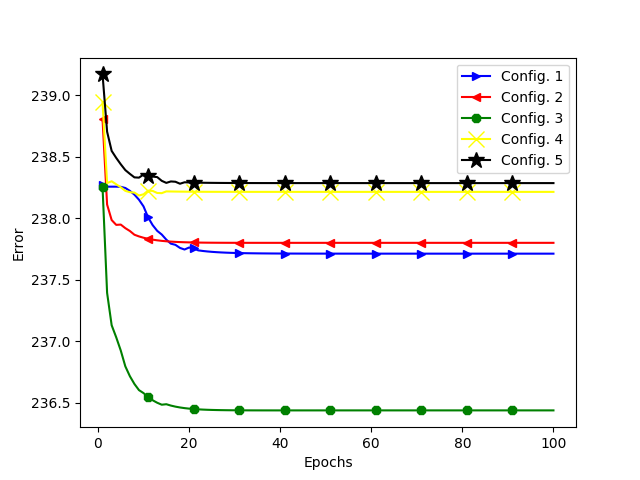}
	\label{Fig_erroconfig}
\end{figure}

The Figure \ref{Fig_erroconfig} shows the error variation in each configuration. It is possible to observe that the error is minimized for all configurations. This error metric is obtained through the sum of all individual sentences errors.

\subsection{Alpha Variation}

\begin{figure}
\centering
	\caption{Variation (interpolated) of alpha.}
	\includegraphics[width=0.7\linewidth]{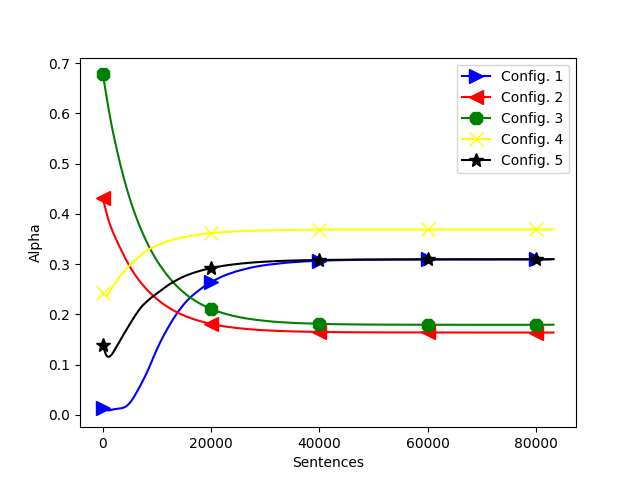}
	\label{Fig_alphaRandom}
	
\end{figure}

Following the same scenario, the Figure \ref{Fig_alphaRandom} illustrates the variation of alpha value over epochs. Note that $\alpha$ lies between $\sim$0.2 and $\sim$0.4. Therefore it can be stated that LSA is more relevant than Naive Bayes, since this value is lower than 0.5.

\subsection{Lambda Optimization}

\begin{figure}
\centering
	\caption{Variation (interpolated) of previous word $\lambda$ about distance 0.}
	\includegraphics[width=0.7\linewidth]{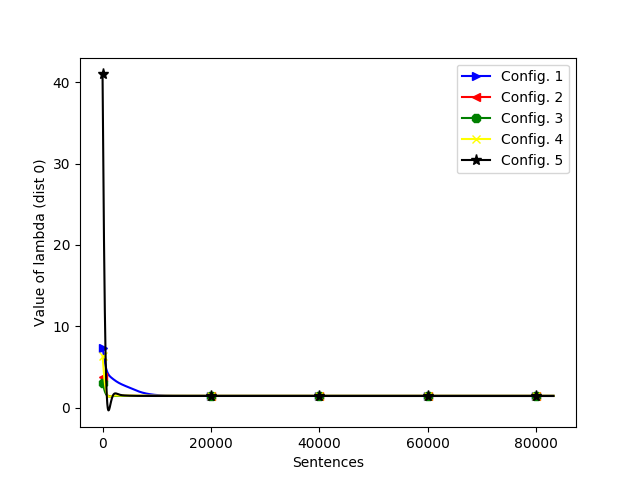}
	\label{Fig_lambPrevious0}
\end{figure}

\begin{figure}
\centering
	\caption{Variation (interpolated) of previous word $\lambda$ about distance 1.}
	\includegraphics[width=0.7\linewidth]{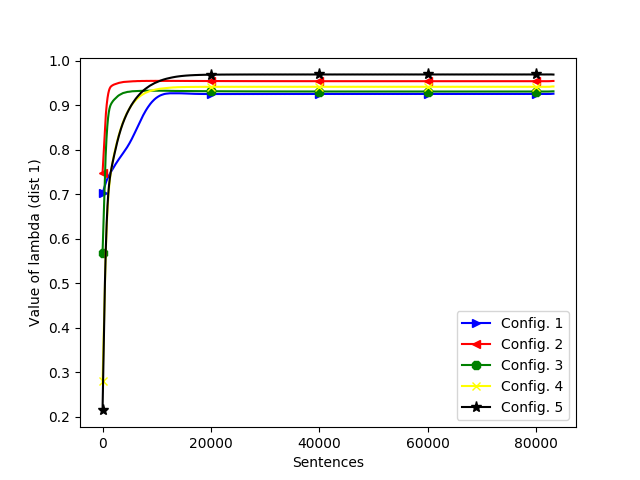}
	\label{Fig_lambPrevious1}
\end{figure}

\begin{figure}
\centering
	\caption{Variation (interpolated) of previous word $\lambda$ about distance 2.}
	\includegraphics[width=0.7\linewidth]{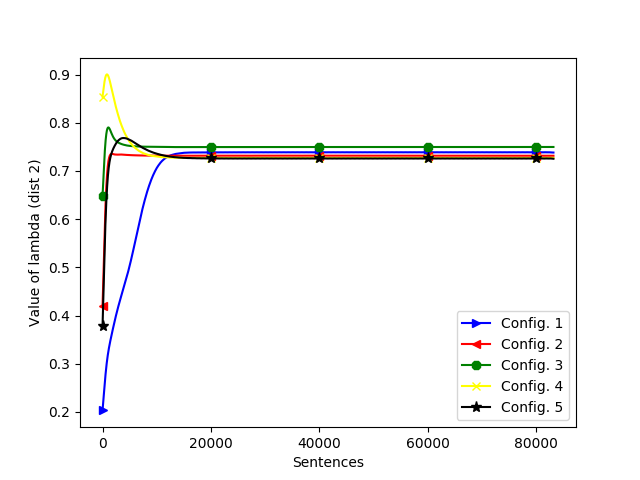}
	\label{Fig_lambPrevious2}
\end{figure}

\begin{figure}
\centering
	\caption{Variation (interpolated) of posterior word $\lambda$ about distance 0.}
	\includegraphics[width=0.7\linewidth]{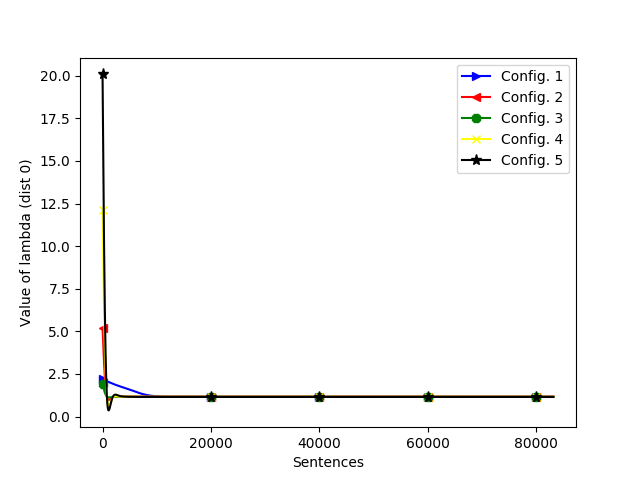}
	\label{Fig_lambPost1}
\end{figure}

\begin{figure}
\centering
	\caption{Variation (interpolated) of posterior word $\lambda$ about distance -1.}
	\includegraphics[width=0.7\linewidth]{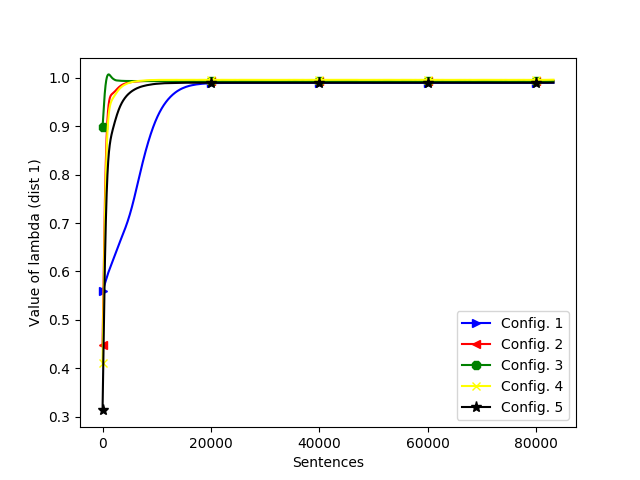}
	\label{Fig_lambPost2}
\end{figure}

\begin{figure}
\centering
	\caption{Variation (interpolated) of posterior word $\lambda$ about distance -2.}
	\includegraphics[width=0.7\linewidth]{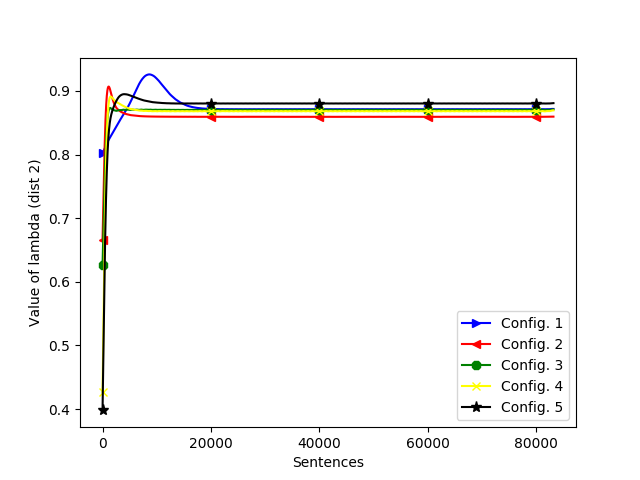}
	\label{Fig_lambPost3}
\end{figure}

In the scenario described above, it is possible to visualize the values of parameters $\lambda$ in each configuration over the optimization sentences, as illustrated in Figures \ref{Fig_lambPrevious0}, \ref{Fig_lambPrevious1}, \ref{Fig_lambPrevious2}, \ref{Fig_lambPost1}, \ref{Fig_lambPost2} and \ref{Fig_lambPost3}.

Results show that $\lambda$ values converge to the same region even with the initial random values and configurations. Therefore, it is feasible to state that in this scenario the word about distance 0 is the most relevant for inference and the words about distance 2 and -2 are the least important for inference.

\subsection{Size of History Influence}

\begin{figure}[h]
\centering
	\caption{Error variation depending on the size of history.}
	\includegraphics[width=0.7\linewidth]{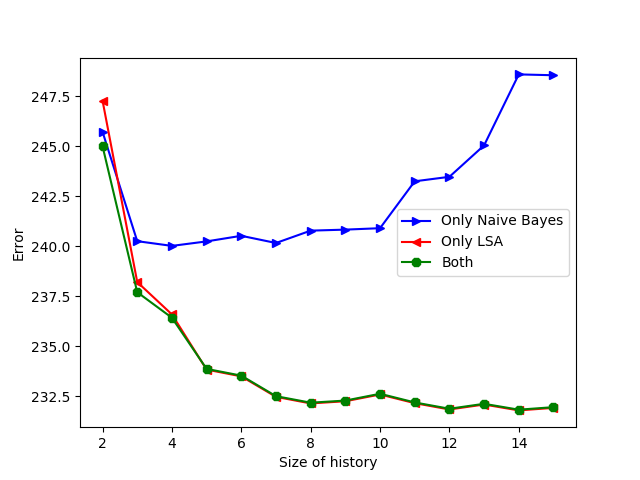}
	\label{Fig_varhist}
	
\end{figure}

In order to understand how the size of history can influence the final result, it was created a scenario where the history was tuned between 2-gram and 15-gram using configuration 1. Furthermore, the proposed model was tested using only Naive Bayes, only LSA and weighing both models (i.e. $\alpha$ optimized), as illustrated in Figure \ref{Fig_varhist}. 

As observed, up to 4-gram there are a weighing between both models that provide better results than these models individually. From that point, LSA provides better results alone. This probably happens because the bayesian model tries to detect patterns which were not present in training.

\subsection{Influence of Each Word for the Naive Bayes}

\begin{figure}
\centering
	\caption{Value of each $\lambda$ in a 15-gram history.}
	\includegraphics[width=1\linewidth]{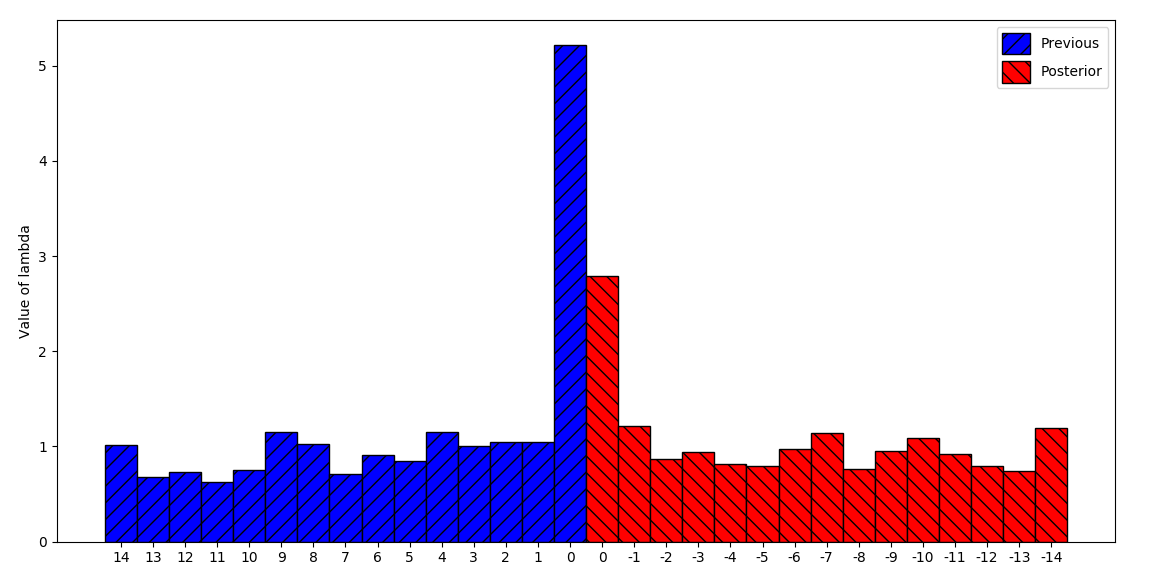}
	\label{Fig_histbayes}
	\vspace{-2mm}
\end{figure}

Through the Naive Bayes it is possible to determine which distances are more or less important for inference, observing the parameters $\lambda$. Thereby, it was created a scenario where a 15-gram history is used and only Naive Bayes is considered (i.e. $\alpha$ = 0) as illustrated in Figure \ref{Fig_histbayes}.
Thereby, both sides of inference have a similar pattern, where the distance 0 words are more relevant.

\subsection{Tests and Comparative}

\begin{figure}
\centering
	\caption{Results of MSR Sentence Completion Challenge and comparative with (A) RNNLM + Skip-gram \cite{mikolov2013efficient}, (B) LSA \cite{zweig2011microsoft} and (C) N-gram \cite{gubbins2013dependency}.}
	\includegraphics[width=0.7\linewidth]{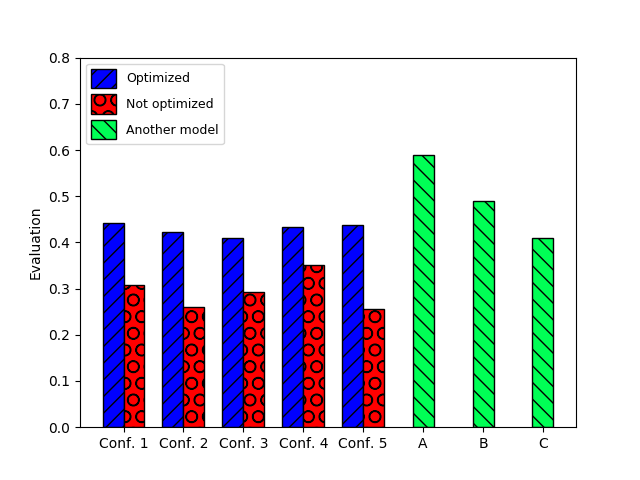}
	\label{Fig_testfull}
\end{figure} 

As explained previously, the proposed model in this work was tested in each configuration using a 3-gram history. The Figure \ref{Fig_testfull} shows the performance of model optimized and non-optimized. Furthermore, it is compared with the state-of-the-art \cite{mikolov2013efficient} and related models.

As observed, the proposed model in this work is superior to another n-gram model which considers history between 5-gram; but it is worse than LSA model, which considers all words in the sentence for inference; also it is worse than neural network model. However, the proposed model provides recommendations considering syntactic and semantic features using less training data than a neural network model.

\section{Conclusion}

In this work, it was developed a hybrid model to complete sentences employing Naive Bayes and LSA models. This hybrid model uses a set words to infer the next word considering each sub-model individually.

In order to validate the proposed hybrid model, it was conducted an experiment in which it is composed by sentences with a missing word. Therefore, there were proposed the parameters $\lambda_1, ..., \lambda_{n-1}$ to optimize the inferences of Naive Bayes and another parameter $\alpha$ to improve their integration with the LSA. To optimize the values of these parameters, it was applied the Gradient Descent technique, where these values are adjusted based on an error function.

Results show that lambda values converge to same values. That is, for each lambda $\lambda$, different experiments result to similar final values. This is a strong 	evidence that the optimization technique is working. Furthermore, results show that the most relevant history information for the suggestion are the previous and posterior neighboring words, as illustrated in Figure \ref{Fig_histbayes}.

Within the optimization experiment, the $\alpha$ parameter was also trained. Results show that $\alpha$ tends to give more importance to LSA model than Naive Bayes, since its value converged to the region between $\sim$0.2 and $\sim$0.4. Another useful result is that the size of history words affects the convergence value of $\alpha$. That is, if the size of history is less than 4, $alpha$ lies between $0$ and $1$. However, if the size of history is higher than 4, $alpha$ tends to $0$, which means that LSA model is the unique model used to suggest the next word. Maybe, this is a evidence that the Naive Bayes model could not find suitable conditional probabilities in the Naive Bayes graph. This behavior could happen if the train database is small and does not contains some relations between words.


One important regard about the experiments is that it was used the cross-validation in order to validate the convergence values. It was found that, for different sets of train samples, the $\lambda$ and $\alpha$ parameters did not change the final optimized value.


During tests, the proposed model proved to be relevant to other related models. Even not providing better results than actual state-of-the-art, the proposed model brought relevant results, considering syntactic/semantic rules among words and consuming a fraction of training required by a model based on Deep Learning.

Thereby, the proposed model in this work achieved the objectives providing a great time $\times$ accuracy relation which can be improved, thus creating new perspectives in NLP as also different applications such: a better word prediction model for mobile platforms, improve the conversation ability of robots, enhance search engines and improve the social interaction of impaired people.

\section{Future Work}

The proposed model has several features which were not investigated. One of these features is to use a bigger database for optimization since the MSR Sentence Completion Challenge has a few supervised examples to do this task.

In this work it was used a fixed number for learning rate $\eta_\alpha$ and $\eta_\lambda$. In case of those number were changed over the epochs, could provide better results.

It was proposed a minimal conditional probability for the Naive Bayes inferences whenever a co-occurrence is not present in the words graph. This gives a minimal chance for a word to be recommended, instead of zeroed product in the bayesian inference. In case of this value is optimized, maybe the bayesian model provides greater results.

The database from Project Gutenberg gives a set of full books (where are included meta-information such as: chapter indicator, paragraphs indicators, publisher notes, etc). In this work, we used a simple regular expression rule to extract useful information from this database. However, much information was lost during this process. It would be better to use text mining algorithms to extract more information in order to get a bigger dataset.

Over the optimization, only the correct option was considered for error minimization. Also, it is possible to consider wrong words to compute the error function. This could bring more precision to the inference process. 

To fairly compare the Naive Bayes and LSA, it was proposed a equalization procedure of output probabilities for each model. This technique worsens the optimization of parameters $\lambda$ when giving normalized values instead of the original. Thereby, another technique that avoids this problem could improve the proposed model.

In Naive Bayes, it was added many parameters $\lambda$ to enhance the bayesian  performance. However, the prior probability does not have one of these parameters. In case of implementing this new $\lambda$ term, it would be possible to observe not only the influence of previous words than posterior but also the prior probability term.

\ifCLASSOPTIONcompsoc
  \section*{Acknowledgments}
\else
  \section*{Acknowledgments}
\fi
The authors would like to thank Telefônica VIVO partnership with Centro Universitário FEI and IIoT Lab Telef\^{o}nica/FEI (Intelligent Internet of Things Lab) by the funding and hardware resources provided.

\bibliographystyle{IEEEtran}
\bibliography{referencias}

\end{document}